\documentclass[10pt,journal,compsoc]{IEEEtran}
\newif\ifpeerreview

\peerreviewfalse

\usepackage[nocompress]{cite}
\usepackage{url}
\usepackage{amsmath,amssymb,graphicx}

\usepackage{lipsum} 

\usepackage[switch]{lineno}

\usepackage{booktabs}
\usepackage{algorithm}
\usepackage{algpseudocode}
\usepackage[position=top]{subfig}
\usepackage{graphicx}

\usepackage{algpseudocode}
\algnewcommand\algorithmicforeach{\textbf{for each}}
\algdef{S}[FOR]{ForEach}[1]{\algorithmicforeach\ #1\ \algorithmicdo}
\newcommand{\paperID}{0034}

\usepackage{tikz}
\def\checkmark{\tikz\fill[scale=0.4](0,.35) -- (.25,0) -- (1,.7) -- (.25,.15) -- cycle;} 

\usepackage{xparse}

\NewDocumentCommand\DownArrow{O{2.0ex} O{black}}{%
   \mathrel{\tikz[baseline] \draw [<-, line width=0.5pt, #2] (0,0) -- ++(0,#1);}
}

\usepackage[pagebackref=true,breaklinks=true,colorlinks,bookmarks=false]{hyperref}

\title{Test-time Training for Hyperspectral Image Super-resolution }

\author{Ke Li,
         Luc~Van~Gool, and Dengxin Dai
\IEEEcompsocitemizethanks{
\IEEEcompsocthanksitem K.~Li is with the Department of Information Technology and Electrical Engineering, ETH Z\"urich, Switzerland.\protect\\
\IEEEcompsocthanksitem L.~Van Gool is with the Department of Information Technology and Electrical Engineering, ETH Z\"urich, Switzerland, with the Department of Electrical Engineering, KU Leuven, Belgium, and with INSAIT, Bulgaria.\\ 
\IEEEcompsocthanksitem D.~Dai is with Huawei Z\"urich Research Center, Switzerland.\protect\\}
}

\begin{document}

\IEEEtitleabstractindextext{%

\begin{abstract}
The progress on Hyperspectral image (HSI) super-resolution (SR) is still lagging behind the research of RGB image SR. HSIs usually have a high number of spectral bands, so accurately modeling spectral band interaction for HSI SR is hard. Also,  training data for HSI SR is hard to obtain so the dataset is usually rather small. 
In this work, we propose a new test-time training method to tackle this problem.
Specifically, a novel self-training framework is developed, where more accurate pseudo-labels and more accurate LR-HR relationships are generated so that the model can be further trained with them to improve performance. In order to better support our test-time training method, we also propose a new network architecture to learn HSI SR without modeling spectral band interaction and propose a new data augmentation method Spectral Mixup to increase the diversity of the training data at test time. 
We also collect a new HSI dataset with a diverse set of images of interesting objects ranging from food to vegetation, to materials, and to general scenes. Extensive experiments on multiple datasets show that our method can improve the performance of pre-trained models significantly after test-time training and outperform competing methods significantly for HSI SR. 

\end{abstract}

}

\maketitle

\section{Introduction}
\label{sec:intro}

Hyperspectral (HS) images contain detailed spectral information of imaged objects, allowing for fine-grained identification and discrimination of materials. They have proven very useful in numerous important applications such as medical diagnosis~\cite{medical:review:14}, food quality and safety control~\cite{food:review:07}, remote sensing~\cite{remote:sensing:09} and object detection~\cite{hyspectraobjdetection}. Given that computer vision technologies have made a series of leaps over the last decade, it is natural to expect a booming literature for automated applications based on HS images as well. This, however, has not happened and probably will not happen soon, due to the lack of affordable cameras to acquire high-resolution HS (HrHS) images efficiently. 

To alleviate this, enormous efforts have been made to design image-processing algorithms to enhance the spatial resolution of low-resolution HS (LrHS) images. A plethora of techniques have been developed to solve this problem \cite{HSISR:fusion:survey:17}. 
Recent research on hyperspectral image super-resolution (HSI SR ) has achieved great success due to the development of deep learning~\cite{pansharpening:CNN16,panet:iccv17,target:cnn:pansharpening,fusion:net:cvpr19,Zheng2020HyperspectralPU, pansharpening:transformer:invertible:22,HyperTransformer}. However, compared to other tasks such as RGB image super-resolution, the methods for HSI SR are still limited to small datasets. The key reason for this is that acquiring large-scale training datasets for supervised HSI SR methods is very challenging because existing HSI cameras to obtain high-resolution HSIs are expensive, bulky, and very slow. 
As a result, the trained models for HIS SR usually do not generalize very well to new and different data samples during testing. 

In this work, we propose test-time training for HSI SR to further adapt the model to new data during testing. In this setting of test-time training, the model has only one input data sample and its own parameters. Since this input sample does not have ground-truth values, the learning is usually done by self-training with synthesized training data.  Therefore, the learning scheme needs to be carefully designed such that the model is able to extract useful information from a single `unlabeled' training sample while not causing the model to diverge from its training.

This work proposes a novel method with three contributions to tackle the main challenge of test-time training for HSI SR. Specifically, the contributions are made to generate effective training data and to have a suitable network architecture for stable learning. The first component of the proposed method aims to improve the quality of the synthesized training sample (i.e. a pair of LR HSI $\mathbf{x}$ and HR HSI $\mathbf{X}$) for the input LR HSI $\mathbf{x}$. We propose to improve it under the self-training framework in two different ways. On the one hand, motivated by the fact that the mean teacher model $\Theta'$ can generate higher-quality predictions than the student model $\Theta$, we use a weight-averaged teacher model to generate the HR HSI prediction $\Bar{\mathbf{X}}$ for the test input image $\mathbf{x}$: $\Bar{\mathbf{X}} = \Theta'(\mathbf{x})$. On the other hand, in order to make sure that the synthesized training sample follows the authentic LR-HR relationship, we downsample the HR HSI prediction $\Bar{\mathbf{x}} = \DownArrow\langle\Bar{\mathbf{X}}\rangle$ and use $(\Bar{\mathbf{x}}, \Bar{\mathbf{X}})$ instead of $(\mathbf{x}, \Bar{\mathbf{X}})$ to form the synthesized training sample for the input image $\mathbf{x}$. The pseudo-sample synthesis and model training reinforce each other as the training process iterates further -- the synthesized training sample is used to fine-tune the given model which is then used to generate a new training sample.  These two simple changes significantly improve the performance of the method compared to standard self-training.

In principle, any deep neural network model can be further trained this way during test time to improve its performance. However, given the fact that only one single training sample is available, one needs to use the right network architecture and carefully select the set of parameters that should be updated during this test-time training. As a second contribution, we propose a simple grouped super-resolution network that fully focuses on spatial image super-resolution. It super-resolves each of the spectral bands separately with the same network. We reason in the method section and also show with experiments that one can achieve very good HSI SR results without modeling spectral band interaction at all.  This design avoids entangling the spectral information of different bands. It not only shows state-of-the-art results for standard HSI SR but also offers a good solution to test-time training for HSI SR. 

Finally, in order to further increase the amount of training data at test time, we propose a novel data augmentation method called Spectral Mixup that mixes the content of different bands to create new spectral images without breaking the detailed structure of HSIs. The data augmentation is specifically designed for HSIs and has shown better performance than other augmentation methods. 


Overall, we make three contributions to enable test-time training for HSI SR. We demonstrate the effectiveness of our proposed approach on three benchmarks, showing that all the proposed contributions are significant and our final approach significantly outperforms other approaches.

\section{Related Work}
\label{sec:formatting}

\noindent
\textbf{Hyperspectral Image Super-Resolution}.
HSI SR has been tackled in multiple different settings: 1) learning HSI SR from only RGB images; 2) learning  HSI SR from low-resolution HSIs (LrHSIs); and 3) learning HSI SR from paired high-resolution RGBIs and LrHSIs of the same scene. Our method falls into the second group. 

HSI recovery from only RGB images is a very challenging problem. However, due to its simple setup, this research direction has gained great traction in recent years~\cite{Arad_2020_CVPR_Workshops}. Regarding the methods used, the general trend is to largely shift from `conventional' methods such as radial basis functions~\cite{Training:Spectral:rgb:14} and sparse coding~\cite{arad_and_ben_shahar_2016_ECCV} to deep learning methods~\cite{Galliani2017LearnedSS,HSCNN+,Arad_2020_CVPR_Workshops}. Due to the grand challenge of reconstructing high-dimensional HSIs from only three channels of RGB images, there emerges research working on recovering HSIs from a larger number of wide spectral bands~\cite{chen2021hyperspectral,Li_2023_WACV}.


Image SR models aim to learn the relationship between the LR images and HR ones by learning from a collection of examples consisting of pairs of HR images and LR images. RGB image super-resolution has achieved remarkable results in the last ten years. The research on using deep neural networks to solve this task started in 2014 \cite{sr:eccv14}. A great deal of progress has been made since then. They focus on making networks deeper or wider~\cite{very:deep:SR:16,residual:dense:sr:cvpr18}, using better feature extractor~\cite{deep:laplacian:cvpr17} and better losses~\cite{Ledig_2017_CVPR}, or modeling providing more realistic image degradation model~\cite{guo2020closed}.  
As to single-image HSI SR, there has been excellent pioneering work~\cite{HSI:SR:05,HIS:SR:2011} as well. In recent years, the classical methods have been outperformed by deep learning methods. For example, a single-band SR is trained on natural image datasets and it is applied to HSIs to mainly address the spatial image super-resolution problem. The spectral band interaction is then separately modeled via matrix factorization afterward. In order to explore both spatial and spectral correlation at the same time, methods based on 3D Convolutional Networks~\cite{3d:net:HSI:sr:17,2d:3d:net:HSI:SR:20} have been developed. Although it sounds intuitive, 3D CNNs are computationally expensive.  To alleviate this, Grouped Convolutions (GCs) with shared parameters have been recently used for this task~\cite{HSI:SR:grouped:recursivenet:18,spatial:spectral:prior:20, Li_2022_WACV}. 


There are cameras that can provide both LrHSIs and HrRGB images. Therefore, quite some research effort has been put into fusing HrRGB images of the same scene as references to improve the spatial resolution of the LrHSIs~\cite{unmixing:survey:12,HSISR:fusion:survey:17, xue2021spatial}.  The main focus of this line of research is to learn to borrow the structure of the HrRGB image and transfer it to the domain of HSIs. While many fusion algorithms have been developed over the past years, they seem to all assume that the RGB image and the HSI image are very well co-registered~\cite{spatial:spectral:prior:20}. Most of the work will just use a completely synthetic setup where images of the two domains are perfectly aligned. This assumption does not hold real-world applications, where data registration is a real challenge and registration errors can lead to degraded SR results~\cite{data:fusion:15,fusion:20}. Our work thus chooses to recover HrHSI from  LrHSI only, and the performance via test-time training. 

\subsection{Test-time Training} 
General domain adaptation requires access to both source and target data. Test-time adaptation methods, however, do not require any data from the source domain. 
There are works that learn an auxiliary task during test time to help align the feature of test data distribution and source data distribution~\cite{li2020model,yeh2021sofa,kurmi2021domain}.

Another major line is to directly fine-tune the pre-trained model on the test samples without explicitly conducting domain alignment. The main research problems are 1) how to regulate the self-training process so that the training process is stable; and 2) what parameters to update to avoid model divergence during training.  
Notable examples include Test entropy minimization~(TENT)~\cite{wang2020tent} which takes a pre-trained model and adapts to the test data by fine-tuning the BN parameters with its own pseudo-labels.  Entropy minimization and a diversity regularizer are then used together for test-time adaptation in Source hypothesis transfer~(SHOT)~\cite{liang2020we}. 
A diversity regularizer and an input transformation module are used to improve the performance in \cite{mummadi2021test}.  In \cite{iwasawa2021test}, the authors choose to only update the final classification layer during inference time using pseudo-prototypes.  
There is a stream of approaches that choose to only update the statistics in the Batch Normalization layer using the target data~\cite{li2016revisiting,hu2021mixnorm,you2021test}. 

While great progress has been made on this topic, most of the approaches still focus on high-level tasks such as image classification~\cite{li2016revisiting,hu2021mixnorm,you2021test}, semantic segmentation~\cite{liu2021source,kundu2021generalize,hu2021fully} or depth estimation \cite{10161304}. Using test-time training for low-level vision tasks has also attracted some research attention recently. For instance, it is used to improve the performance of RGB image super-resolution by fine-tuning the model on a few reference images at test time \cite{9607414}. Their setting and tasks are very different from ours. Very recently, Deng \emph{et. al.}  \cite{deng2023efficient} have proposed a test-time domain adaptation method for RGB image super-resolution, which is able to quickly adapt SR models to test domains with different/unknown degradation types.


\section{Test-time Training for HSI SR}

While there is still no camera that can record HrHS images efficiently, cameras for a compromised setting to obtain LrHS images are becoming common. Due to this reason, HSI SR models are trained to increase the resolution of low-resolution HSIs. Given a  dataset $\mathcal{D} = \{\mathbf{x}_n,\mathbf{X}_n \}_{n=1}^N$, where $\mathbf{X}_n \in \mathbb{R}^{HW\times S}$ is the desired HrHS image with its size $H\times W$ and band number $S$, and $\mathbf{x}_n \in \mathbb{R}^{hw\times S}$ is the LrHS image with its size $h\times w$ and band number $S$, a deep neural network $\Theta$ is trained to approximate the mapping function $\Theta(\mathbf{X}|\mathbf{x})$. The trained model can then be applied to new LrHSIs.  The goal of this work is to further train this model during test time and improve its performance for the given test sample. This can be very useful when the test sample is from a different data distribution such as a different type of scene or when the source training dataset $\mathcal{D}$ is quite small. 

Given a trained model $\Theta_0(\mathbf{x})$ trained on the source data $\mathcal{D}$, we aim at improving the performance of this existing model for a test sample $\mathbf{x}_t$ during testing time. We do this via test-time training without access to any source data. 
We propose an adaptation method for this test-time training setup. The proposed method takes an off-the-shelf source pre-trained model and adapts it to the given test sample via a new self-training method. 
The new method is proposed due to two reasons. First, the LR-HR relationship between the LR input (i.e. the input test sample) and the HR input (i.e. the HR prediction of the input sample) is not completely accurate; self-training will reinforce and accumulate this error. Second, running self-training with a single training sample can lead to model divergence as the training process iterates. The new approach consists of three contributions: 1) Introducing a simple yet effective HSI SR network (i.e. the source model) to improve the training stability in this low data regime; 2) Developing a novel training sample generation approach to generate both accurate pseudo HR ground truth and accurate LR-HR relationship; and 3) Proposing a new data augmentation method called spectral mixup to generate more training samples which further boosts the performance of the method.

\subsection{Self-training with Pseudo Samples}
\label{sec:weight-averaged} 

Our self-training iterates over $T$ steps in total. At each step, it generates a pair of HrHSI and LrHSI as the training data and then fine-tunes the model with this generated sample. This process continues until $T$ steps are achieved. 
Given target data $\mathbf{x}$ and the model $\Theta_t$, the common test-time objective under the self-training framework would be to minimize the inconsistency between the prediction and the pseudo-label. Directly using the model's prediction (after label sharpening) as the pseudo-label leads to the training objective of TENT~\cite{wang2020tent}~(i.e. entropy minimization) for image classification. While this works well for semantic recognition tasks,  it does not work for a regression task such as SR as training $\Theta_t$ with training sample $(\mathbf{x}, \Theta_t(\mathbf{x}))$ will lead to zero updates. 

In order to address this problem, we introduce two modifications to the standard self-training making it work well for HSI SR.
First, we propose to use $(\DownArrow\langle\Theta_t(\mathbf{x})\rangle, \Theta_t(\mathbf{x}))$ instead of $(\mathbf{x}, \Theta_t(\mathbf{x}))$ to form the synthesized training sample for test-time training to improve the performance for input $\mathbf{x}$, where $\DownArrow\langle\mathbf{X}\rangle$ is a downsampling operator to downsample $\mathbf{X}$ to the size of the test sample $\mathbf{x}$.
The model $\Theta_t$ is then updated by gradient descent using the following loss:
\begin{equation}
\mathcal{L}_t =  \mathcal{L}(\Theta_t(\mathbf{x}), \Theta_t(\DownArrow\langle\Theta_t(\mathbf{x})\rangle))
\label{eq:pseudo}
\end{equation}
Basically, for the new training sample, the LR-HR relationship of $(\DownArrow\langle\Theta_t(\mathbf{x})\rangle,\Theta_t(\mathbf{x}))$ is perfectly accurate so that the model is able to learn useful information for HSI SR for the actual input  $\textbf{x}$ from a training sample $\DownArrow\langle\Theta_t(\mathbf{x})\rangle$ which is very similar to $\textbf{x}$. Because they are very similar, the updated model can be easily `generalized' to  $\textbf{x}$ in the following step $t+1$ to get a better $\Theta_{t+1}(\mathbf{x})$ such that it can generate even better training data. The data generation and model training reinforce each other as the training process iterates. 


Second, motivated by the observation that weight-averaged models over training steps can provide a more accurate model than the student model~\cite{polyak1992acceleration, tarvainen2017mean,Wang_2022_CVPR}, we use a weight-averaged teacher model $\Theta_t'$ to generate the pseudo-labels.  At iteration step $t=0$, the teacher network is initialized to be the same as the source pre-trained network. At iteration step $t$, the pseudo-label is first generated by the teacher $\Theta_t'(\mathbf{x})$. The student model $\Theta_t$ is then updated by gradient descent using the consistency loss between the student's and teacher's predictions:
\begin{equation}
\mathcal{L}_t =  \mathcal{L}(\Theta_t'(\mathbf{x}), \Theta_t(\DownArrow\langle\Theta_t(\mathbf{x})\rangle)).
\label{eq:weight_consistency}
\end{equation}

The loss enforces consistency between the teacher and student predictions and can guide the training of the student model. 

After the update of the student model $\Theta_t \xrightarrow{} \Theta_{t+1}$ using Equation~\ref{eq:weight_consistency}, we update the weights of the teacher model by exponential moving average using the student weights: 
\begin{equation}
\Theta'_{t+1} =  \alpha \Theta'_{t} + (1-\alpha) \Theta_{t+1}, \label{ema}
\end{equation}
where $\alpha$ is a smoothing factor.  

As also pointed out by~\cite{wang2020tent}, the benefits of the weight-averaged consistency are two-fold: 1) by using the often more accurate weight-averaged prediction as the pseudo-label target, our HSI SR model is able to update itself for test-time training; and 2) the mean teacher prediction encodes the information from models in past iterations and is, therefore, less likely to suffer from catastrophic forgetting over the learning iterations. The weight-averaged pseudo-label is largely inspired by the mean teacher method proposed in~\cite{tarvainen2017mean} in the semi-supervised learning setup.

\subsection{Source HSI SR Model}
In this section, we present the architecture of our HSI SR network $\Theta$. Feeding all channels of the low-resolution HSI (LrHSI) to a standard network and letting it output all channels of the upsampled high-resolution HSI (HrHSI) would be a straightforward choice. However, this choice is actually not ideal for HSI SR. For HSI SR, the number of channels $S$ is usually high (tens to hundreds), and dataset size $N$ is usually not very large (at the scale of a few hundred). Intuitively, processing all the channels together through one neural network provides the opportunity to let all the spectral bands fully interact with each other such that they provide useful information to each other to generate higher-quality SR outputs. While it is intuitive, there is actually one drawback to this choice that has not received research attention. This choice can easily make all the spectral bands deeply entangled and a great deal of learning effort is then needed to disentangle them again. In other words, the network is not fully focused on SR itself because it also needs to learn how to disentangle the entangled spectral bands. As an alternative, we propose to train an HSI SR network that is fully focused on spatial image super-resolution without modeling any spectral band interaction at all. This new route may sound counter-intuitive as it loses the opportunity to model spectral band interaction. However, the benefit of avoiding entangling the information of all the bands and avoiding the risk of not completely disentangling them outweighs the loss of the presumed capability of modeling spectral band interaction. This new choice improves the performance of standard HSI SR. More importantly, it is especially suitable for test-time training of HSI SR because this new design is more compact and thus less prone to training instability when training with a very small amount of training data.  


To have the flexibility of arbitrary-scale upsampling, we employ the Local Implicit Image Function \cite{implicit} as the basis to build our spatial image super-resolution method. For a single channel image $\mathbf{x}^s$ with $s \in [1, S]$, we use a standard deep neural network to extract a dense feature map $\mathrm{Z}^s$. We then use the local image implicit function to decode the feature map $\mathrm{Z}^s$ to obtain a dense image $\bar{\mathrm{X}}^s$.  Following \cite{implicit}, we parameterize the decoding function $f_\theta$ as an MLP that takes the form: 
\begin{equation}
    \bar{\mathrm{X}}^s(n) = f_\theta(\mathbf{z}^s(n^\prime),n-n^\prime),
    \label{eq:implicit1}
\end{equation}
where $\mathbf{z}^s(n^\prime)$ is the nearest latent code from $n$ in $\mathrm{Z}^s$.

The idea is that a continuous image is represented as a 2D feature map $\mathrm{Z}^s \in \mathbb{R}^{H\times W\times D}$, where $D$ is the feature dimension.  This function $f_\theta$ is shared by all the images. 

As pointed out by \cite{implicit}, direct use of Equation \ref{eq:implicit1} can lead to discontinuous predictions for the `border' pixels where the selection of the nearest latent code $\mathbf{z}^s(n^\prime)$  switches. We follow the general idea of \cite{implicit} and address this by using a local ensemble so that Equation \ref{eq:implicit1} is extended to 
\begin{equation}
        \bar{\mathrm{X}}^s(n) = \frac{\sum_{t=1,2,3,4}\frac{1}{\|n-n^\prime_t\|^2}.f_\theta(\mathbf{z}^s(n^\prime_t),n-n^\prime_t)}{\sum_{t=1,2,3,4}\frac{1}{\|n-n^\prime_t\|^2}},
    \label{eq:implicit2}
\end{equation}
where $\mathbf{z}^s(n_t^\prime)$ ($t \in \{1,2,3,4\}$) are the 4 nearest latent codes for query location $n$.

By now, we have an SR method for a single spectral band. Another challenge facing us before delivering a full HSI SR algorithm is to handle $S$ spectral bands. 
To address this, we use the idea of grouped convolutions with group size $1$, i.e. each band is a single group. This means that we share the same feature encoding network and the same dense image decoding network across all bands. We generate the upsampled images $\bar{\mathrm{X}}^s, s = [0,1,...,S]$ for all bands, and then concatenate them to form the final $S$-channel upsampled image $\bar{\mathrm{X}} \in \mathbb{R}^{H\times W \times S}$.  We now have a complete method for our HSI SR, which is compact and works well without modeling spectral band interaction.

\textbf{Loss}. In order to capture both spatial and spectral correlation of the SR results, we follow \cite{spatial:spectral:prior:20} and combine the L1 loss and the spatial-spectral total variation (SSTV) loss \cite{sstV:tip:16}. SSTV is used to encourage smooth results in both spatial domain and spectral domain and it is defined as: 
\begin{equation}
    \mathcal{L_{\text{SSTV}}}=\frac{1}{N} \sum_{n=1}^N (||\triangledown_{\text{h}}\hat{\mathrm{X}}^n||_1 + ||\triangledown_{\text{w}}\hat{\mathrm{X}}^n||_1 + ||\triangledown_{\text{c}}\hat{\mathrm{X}}^n||_1),
\end{equation}
where $\triangledown_{\text{h}}$, $\triangledown_{\text{w}}$, and $\triangledown_{\text{c}}$ compute gradient along the horizontal, vertical and spectral directions, resp. The loss is: 
\begin{equation}
    \mathcal{L}= \mathcal{L}_1 +  \mathcal{L_{\text{SSTV}}}.
\end{equation}

\subsection{Spectral Mixup} 
\label{sec:spectralmixup}
Data augmentation is a strategy to create virtual samples by adding perturbations to the original samples. It is especially useful when the amount of training data is limited. This is exactly the case for our test-time training task. Therefore, we also propose a new data augmentation method to further improve the performance of our test-time adaptation method.
The \emph{mixup} method~\cite{zhang2018mixup} and other image perturbation methods such as color jittering are effective for high-level recognition tasks. However, they are not very useful for low-level SR tasks \cite{data:augmentation:SR:cvpr20} because the detailed image structures can be destroyed by many of those operations. These detailed structures are important for SR tasks, especially for HSI SR. Taking into account this observation, we propose a data augmentation method \emph{Spectral Mixup} specifically for HSI SR. It creates augmented HR HSI samples by using convex combinations of the spectral bands of a given HR HSI sample, and then downsample to LR images.

\begin{algorithm}[t]
\caption{The proposed test-time training approach}
\label{algo}
\textbf{Initialization: } A source pre-trained model $\Theta_0(\mathbf{x})$, teacher model $\Theta'_0(\mathbf{x})$ initialized to $\Theta_0(\mathbf{x})$. \\
\textbf{Input: } the LR HSI $\mathbf{x}$. 
\begin{algorithmic}[1]
\ForEach {$t \in [1,T] $}:
\State Get the HR prediction by teacher $\Theta'_t(\mathbf{x})$.
\State Down-sample HR prediction  $\DownArrow\langle\Theta'_t(\mathbf{x})\rangle$.
\State Update student $\Theta_t$ according to Equation~\ref{eq:weight_consistency}.
\State Update teacher $\Theta'_t$ according to Equation~\ref{ema}.
\State Augment $\Theta'_t(\mathbf{x})$ by  
Equation~\ref{eq:aug} to obtain $\mathring{\mathbf{X}}$.
\State Down-sample  $\mathring{\mathbf{X}}$ to obtain  $\mathring{\mathbf{x}'}$.
\State Update student $\Theta_t$ according to $\mathcal{L}(\mathring{\mathbf{X}}, \Theta_t(\mathring{\mathbf{x}}))$.
\State Update teacher $\Theta'_t$ according to Equation~\ref{ema}.
\EndFor
\end{algorithmic}
\textbf{Output:} Prediction $\Theta'_T(\mathbf{x})$; Final student model $\Theta_T$; Final teacher model $\Theta'_T$.
\end{algorithm}

More specifically, given an HR HSI $\mathbf{X}$, with $S$ channels, we generate a mixing matrix $\mathbf{B} \in \mathbb{R}^{S \times S}$ with values randomly sampled from a uniform distribution over the interval $[0, 1)$. $\mathbf{B}$ is then row-wise normalized to make sure that the values in the projected image have the same magnitude as that of the original image. The augmented HR HSI is then created as: 
\begin{equation}
     \mathring{\mathbf{X}} = \lambda \mathbf{X} + (1-\lambda )\mathbf{B} \mathbf{X}, 
     \label{eq:aug}
\end{equation}
and the corresponding LR HSI is obtained by downsampling $\mathring{\mathbf{X}}$: $ \mathring{\mathbf{x}} =  \DownArrow\langle\mathring{\mathbf{X}}\langle$. 
The randomly projected images are fused with the original images to strike a balance between increasing variations and preserving the fidelity of real HSIs. The fidelity of the real HSIs is partially kept. In this work, $\alpha$ is set to $0.5$. The implementation of \emph{Spectral Mixup} data augmentation is very straightforward and can be done with a few lines of code. \emph{Spectral Mixup} also introduces very little computation overhead. By applying it, more examples from the vicinity of the original example can be sampled. 

\subsection{Full Algorithm} 

As shown in Algorithm~\ref{algo}, combining the new source HSI SR model, the new self-training method with pseudo samples, and the new spectral mixup data augmentation method, we have our complete approach. The source model is further trained for each test sample $\mathbf{x}$ specifically and its performance is measured exactly on the same test sample $\mathbf{x}$. The idea is to improve the performance of the model on $\mathbf{x}$ after test-time training with this sample.

\begin{figure*}[!tb]
    \centering
    \includegraphics[width=\textwidth]{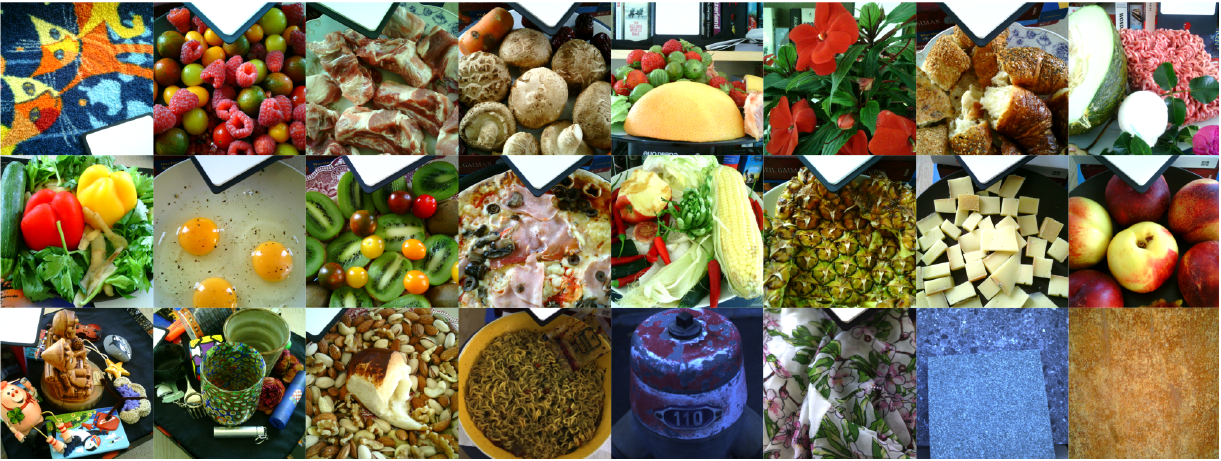}
    \caption{Exemplar samples from our SpectralWorld Dataset.}
    \label{fig:my_label}
\end{figure*}

\section{Experiments}
\label{sec:experiments}

\subsection{Experimental Setup}
\textbf{Datasets}.
We evaluate our method on three datasets. The datasets considered are three nature HSI datasets: CAVE dataset~\cite{cave:10}, Harvard dataest~\cite{harward:dataset}, and a new dataset Spectral World that we collected.  Images in the CAVE dataset have $31$ bands ranging from $400$ nm to $700$ nm at a step of $10$ nm. Images in the Harvard dataset contain $31$ bands as well but range from $420$ nm to $720$ nm. 

The CAVE dataset contains $32$ images of 512 x 512 pixels. We use $20$ images for training and $10$ images for testing. For the Harvard dataset, there are $50$ images in total. We use $40$ for training and $10$ for test.  The Spectral World dataset is collected with a Specim IQ
mobile hyperspectral camera, containing 216 images of 512 × 512 pixels with 204 spectral bands in the 400-1000 nm range. All images were captured using a tripod and by ensuring minimal movement in the scene. All the images have been white-balanced by using a reference white patch to remove the color cast of the illumination.   
The captured dataset includes images of both indoor and
outdoor scenes featuring a diversity of objects, materials, and scales (see Fig. \ref{fig:my_label} for a few example images shown in sRGB). We believe the database to be a representative
sample of real-world images, capturing both pixel-level material statistics and spatial interactions induced by texture
and lighting effects. In addition to the analysis in this paper, these images may be useful for designing and evaluating other HSI research.

\textbf{Evaluation Metrics}.
We follow the literature and evaluate the performance of all methods under three standard metrics \cite{spatial:spectral:prior:20}. They are root mean squared error (RMSE), erreur relative globale adimensionnelle de synthese (ERGAS) \cite{ERGAS:02}, and peak signal-to-noise ratio (PSNR). For PSNR of the reconstructed HSIs, their mean values of all spectral bands are reported as MPSNR. ERGAS are widely used in HSI fusion task.

\textbf{Training Details}. 
We use ADAM optimizer \cite{adam} and train all variants of our method for $T=20$ epochs for each test image. This is still quite fast as we only have one training sample. We find that $20$ epochs are sufficient to give good results for our method -- a larger number probably can further push the numbers up but the difference becomes small. The initial learning rate of all our methods is set to $10^{-5}$. Following previous works in the literature, we set 
$\alpha$ to $0.99$ for all our experiments. 
For the local implicit image function, we use the EDSR-baseline \cite{Lim_2017_CVPR_Workshops}  backbone.

\subsection{Experimental Results}

\begin{table*}[!tb]
  \centering
  \setlength\tabcolsep{1.8pt}
  \small
  \begin{tabular}{l|ccc|ccc|ccccc}
\toprule
 & \multicolumn{3} {|c} {CAVE} & \multicolumn{3} {|c} {Harvard} &\multicolumn{3} {|c} {Spectral World}\\
Methods  & RMSE $\downarrow$ & MPSNR $\uparrow$& ERGAS $\downarrow$ & RMSE $\downarrow$ & MPSNR$\uparrow$& ERGAS $\downarrow$ & RMSE $\downarrow$ & MPSNR$\uparrow$& ERGAS $\downarrow$\\ \midrule
LIIF \cite{implicit}   & 0.0185 & 38.7380 & 5.2719 & 0.0167 &38.8975 & 3.8069 &0.0272 &38.9754 & 1.4945 \\ 
LIIF single-channel (ours)   & 0.0144 & \textbf{39.8385} & \textbf{4.8345} & \textbf{0.0151} & 40.1366 & \textbf{3.3712}  & \textbf{0.0245} & \textbf{40.2182} & \textbf{1.4238}\\
\midrule
LIIF single-channel + SARM~\cite{spatial:spectral:prior:20} & \textbf{0.0142} & 39.4385 & 4.9211 & 0.0158 & 40.0974 & 3.3892  & 0.0249 &40.2014 & 1.4324\\ 
LIIF single-channel + Restormer \cite{Zamir2021Restormer}   & 0.0151 & 38.8234 & 4.9145 & 0.0150 & \textbf{40.2066} & 3.3744  &0.0251 &40.2231 & 1.4217\\ 
\bottomrule 
\end{tabular} 
\caption{Results of all methods on the CAVE, Harvard, and Spectral World datasets for the $\times 2$ case. All methods in this table are source models, without having any test-time training.} 
  \label{tab:1}
\end{table*} 

\begin{table*}[!tb]
  \centering
  \setlength\tabcolsep{1.8pt}
  \small
  \begin{tabular}{cccc|ccc|ccc|ccccc}
\toprule
 \multicolumn{4}{c} {Method} & \multicolumn{3} {|c} {CAVE} & \multicolumn{3} {|c} {Harvard} &\multicolumn{3} {|c} {Spectral World}\\
Source & Pseudo &  Teacher & Mixup  & RMSE $\downarrow$ & MPSNR $\uparrow$& ERGAS $\downarrow$ & RMSE $\downarrow$ & MPSNR$\uparrow$& ERGAS $\downarrow$ & RMSE $\downarrow$ & MPSNR$\uparrow$& ERGAS $\downarrow$\\ \midrule
\checkmark & &&  &  {0.0144} & {39.8385} & {4.8345} & {0.0151} &  {40.1366} & {3.3712}  & {0.0245} & {40.2182} & {1.4238}\\ 
\checkmark & \checkmark &&  & {0.0132} & {40.8385} & {4.7342} & {0.0140} &  {40.8256} & {3.2643}  & {0.0230} & {40.9412} & {1.3296}\\ 
\checkmark &\checkmark & \checkmark &   & 0.0127 & 41.2380 & {4.3782} & {0.0124} &  {41.8720} & {3.0432}  & 0.0214 &41.2901 & 1.2136\\ 
\checkmark & \checkmark & \checkmark & \checkmark    &  \textbf{0.0119} & \textbf{41.8654} &  \textbf{4.2564} &  \textbf{0.0113} &   \textbf{42.3580} &  \textbf{2.9243}  & \textbf{0.0193} & \textbf{41.8038} & \textbf{1.1463}\\ 
\midrule
 \multicolumn{4}{c|} {TTT \cite{sun19ttt}}   & 0.0187 & 38.9843 & 5.0542 & 0.0156 &38.9324 & 3.7906 & 0.0240 &34.8423 & 1.3953\\ 
  \multicolumn{4}{c|} {TTT-MAE \cite{gandelsman2022testtime}}  & 0.0180 & 39.7524 & 4.9342 & 0.0149 &39.2317 & 3.6034 & 0.0227 &37.0923 & 1.2823\\ 
  \multicolumn{4}{c|} {TTAPS \cite{Bartler2022TTAPSTA}}   & 0.0176& 40.0023 & 4.6982 & 0.0142 &39.9543 & 3.5812 & 0.0217 &38.5672 & 1.2543\\ 
\bottomrule
\end{tabular} 
\caption{Results of all methods on the CAVE, Harvard, and Spectral World datasets for the $\times 2$ case. `Source' means the pre-trained source model, `Pseudo' means the generated Pseudo Sample in Eq. \ref{eq:pseudo}, `Teacher' stands for the teacher-student framework, and `Mixup' means the data augmentation method Spectral Mixup. } 
  \label{tab:2}
\end{table*}

\begin{table*}[!tb]
  \centering
  \setlength\tabcolsep{1.8pt}
  \small
  \begin{tabular}{cccc|ccc|ccc|ccccc}
\toprule
 \multicolumn{4}{c} {Method} & \multicolumn{3} {c} {CAVE} & \multicolumn{3} {c} {Harvard} &\multicolumn{3} {c} {Spectral World}\\
Source & Pseudo &  Teacher  & Mixup  & RMSE $\downarrow$ & MPSNR $\uparrow$& ERGAS $\downarrow$ & RMSE $\downarrow$ & MPSNR$\uparrow$& ERGAS $\downarrow$ & RMSE $\downarrow$ & MPSNR$\uparrow$& ERGAS $\downarrow$\\ \midrule
\checkmark & &&  &{0.0162} & {37.7645} & {5.3346} &{0.0167} & {37.9317} & {4.2128} & {0.0150} & {38.9642} &{2.0654}\\  
\checkmark & \checkmark &&   & {0.0152} & {38.6754} & {5.0127} &{0.0153} & {38.6717} & {3.9432} & {0.0135} & {39.7548} &{1.9841}\\ 
\checkmark &\checkmark & \checkmark &   & {0.0145} & {38.9541} & {4.8563} &{0.0145} & {38.9932} & {3.8023} & {0.0129} & {39.9941} &{1.9782}\\ 
\checkmark & \checkmark & \checkmark & \checkmark   & \textbf{0.0139} &  \textbf{39.2376} &  \textbf{4.8001} & \textbf{0.0140} &  \textbf{39.4467} &  \textbf{3.7176} &  \textbf{0.0122} &  \textbf{40.2083} & \textbf{1.8990}\\ 
\midrule
 \multicolumn{4}{c|} {TTT \cite{sun19ttt}}  &{0.0171} & {36.3452} & {5.3788} &{0.0178} & {36.2369} & {4.2934} & {0.0159} & {38.2137} &{2.1371}\\ 
\multicolumn{4}{c|} {TTT-MAE \cite{gandelsman2022testtime}}  &0.0164 & 36.8734 & 5.2144 &0.0167 &36.6754 & 4.2321 & 0.0153 & 38.4653 &2.0933\\ 
  \multicolumn{4}{c|} {TTAPS \cite{Bartler2022TTAPSTA}}  &0.0159 & 37.0125 & 5.1458 &0.0162 & 36.9372 & 4.1784 & 0.0148 & 38.7791&1.9994\\ 
\bottomrule
\end{tabular} 
\caption{Results of all methods on the CAVE, Harvard, and Spectral World datasets for the $\times 4$ case. `Source' means the pre-trained source model, `Pseudo' means the generated Pseudo Sample in Eq. \ref{eq:pseudo}, `Teacher' stands for the teacher-student framework, and `Mixup' means the data augmentation method Spectral Mixup.} 
  \label{tab:3}
\end{table*} 

\textbf{Network architecture comparison}. 
In this section, we compare our network architecture with similar network architectures and show that tacking HSI SR as a single-channel image super-resolution without modeling spectral band interaction is actually a good choice. This comparison is done for the standard HSI SR task, i.e. training on the training dataset and test on the test dataset without test-time training. We employ the same LIIF \cite{implicit} with the same EDSR-baseline \cite{Lim_2017_CVPR_Workshops}  backbone but compare two variants: 1) the standard network architecture where all spectral bands are used as input and let it predict all spectral bands as outputs; and 2) our proposal where one single spectral band is used as input and let it only predict one single spectral band as the output -- the same network is shared by all the spectral bands and the full HSI is obtained by predicting all the bands one by one.    
The results on the top panel of Table \ref{tab:1} show that learning to upsample each band separately outperforms learning to upsample them together. 

The choice of joint upsampling can easily entangle information of all the spectral bands. A great deal of learning effort is then needed to disentangle them, distracting the network from focusing on solving spatial image SR itself. This conclusion can also be verified by the fact that the difference between the two approaches becomes bigger when the number of spectral bands increases. For instance, modeling each band separately shows more advantages on the Spectral World dataset than on the other two datasets. This is because when the number of bands increases, disentangling all the bands from a common feature representation becomes harder. 

To further investigate this, we attach two networks that are popular for modeling channel interactions to our single-channel network to see whether they can further improve the HSI SR performance. The two networks are Restormer \cite{Zamir2021Restormer} and the Spectral Attention Residual Module (SARM) of SSPSR~\cite{spatial:spectral:prior:20}. Experimental results on the bottom panel of Table \ref{tab:1} show that they do not provide further improvement than our single-channel spatial SR method. This implies that at the current development stage of HSI SR, it is promising to just focus on improving the spatial enhancing method for single-channel images to process each channel individually. Effectively modeling spectral band interaction among a large number of bands probably requires either a much larger dataset to learn or a smarter learning method to do it. Unfortunately, it seems that we do not have either of them. We hope that this study also provides a strong baseline for the community to develop more effective spectral band modeling methods in the future.

\textbf{Test-time Adaptation}. 
In this section, we verify the effectiveness of our test-time training method, compare it with other competing methods, and study the contribution of its components. The results are shown in Table \ref{tab:2} and Table \ref{tab:3} for scaling factor $\times 2$ and $\times 4$, respectively.

\begin{figure*}[!tb]
    \centering
    \subfloat[Ground Truth]{\includegraphics[width=0.245\textwidth]{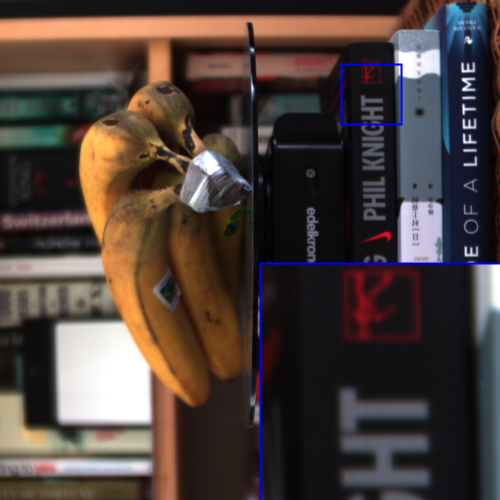}\label{fig:filter1}}
    \hfil
       \subfloat[Bilinear Interpolation]{\includegraphics[width=0.245\textwidth]{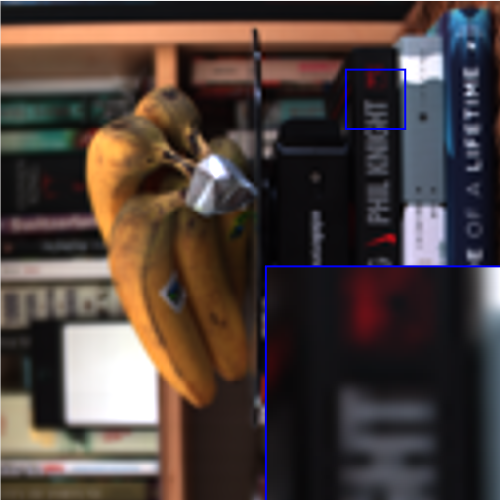}\label{fig:filter2}} 
            \hfil
     \subfloat[Source Model \cite{implicit}]{\includegraphics[width=0.245\textwidth]{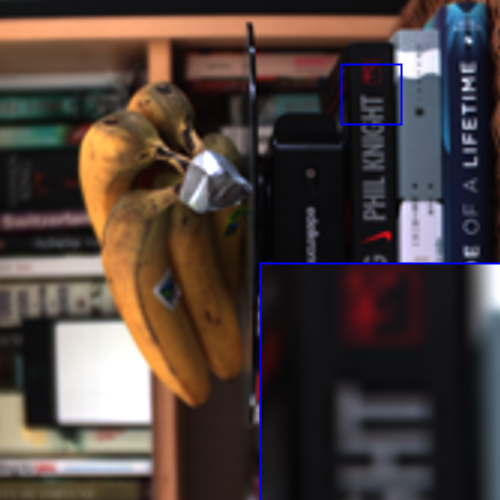}\label{fig:filter3}} 
        \hfil
      \subfloat[Test-time Trained (Ours)]{\includegraphics[width=0.245\textwidth]{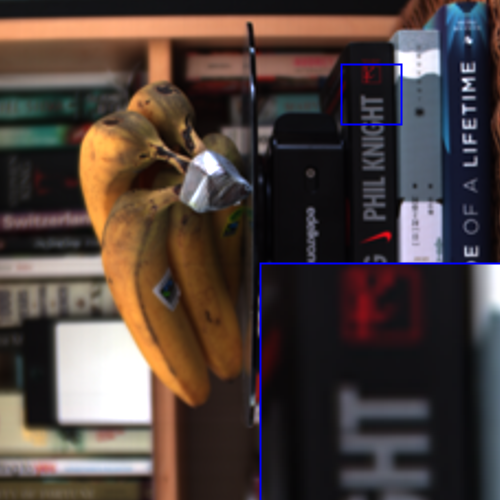}\label{fig:filter3}}  \\ \vspace{-3mm}
    \subfloat{\includegraphics[width=0.245\textwidth]{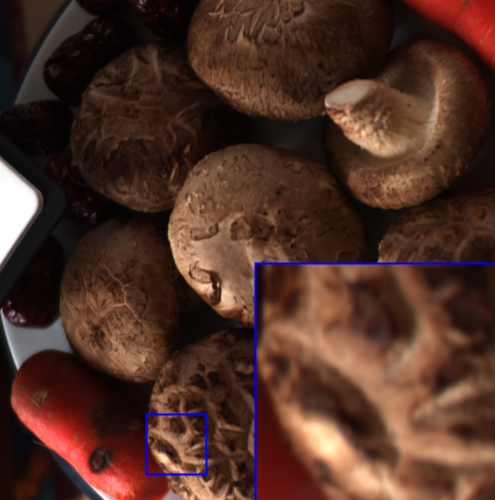}\label{fig:filter1}}
    \hfil
       \subfloat{\includegraphics[width=0.245\textwidth]{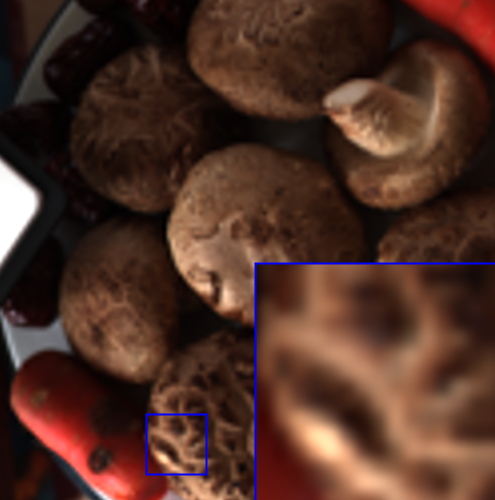}\label{fig:filter2}} 
            \hfil
     \subfloat {\includegraphics[width=0.245\textwidth]{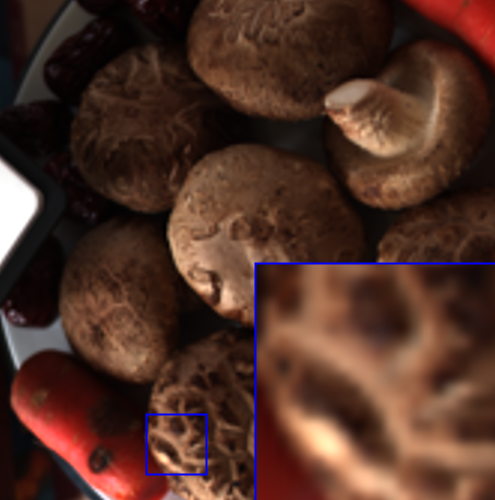}\label{fig:filter3}} 
        \hfil      \subfloat{\includegraphics[width=0.245\textwidth]{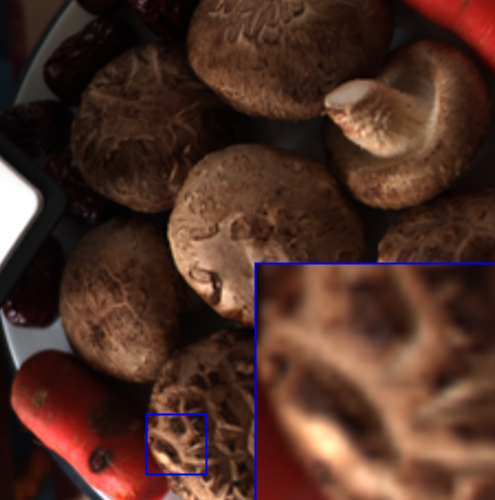}\label{fig:filter3}}  \\ \vspace{-3mm}
        \subfloat{\includegraphics[width=0.245\textwidth]{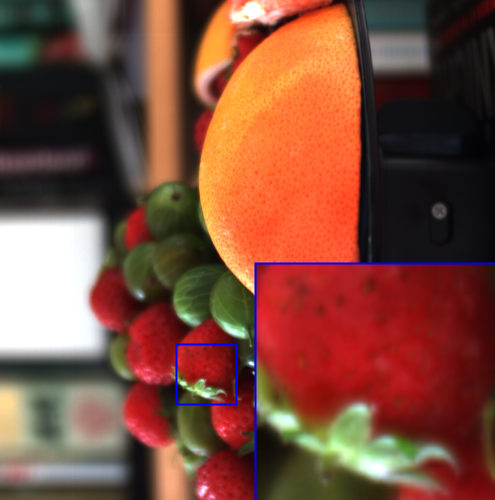}\label{fig:filter1}}
    \hfil
       \subfloat{\includegraphics[width=0.245\textwidth]{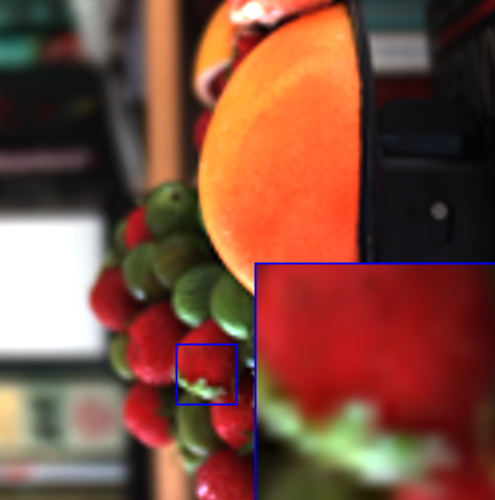}\label{fig:filter2}} 
            \hfil
     \subfloat{\includegraphics[width=0.245\textwidth]{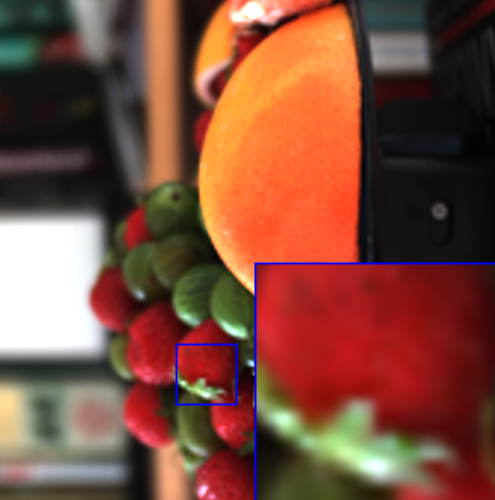}\label{fig:filter3}} 
        \hfil
      \subfloat{\includegraphics[width=0.245\textwidth]{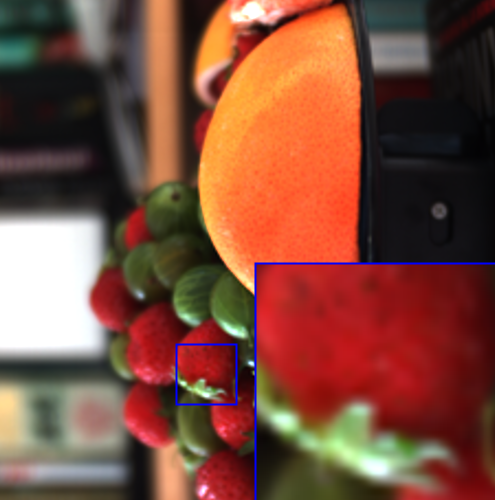}\label{fig:filter3}}  \\
      \vspace{-3mm}
        \subfloat{\includegraphics[width=0.245\textwidth]{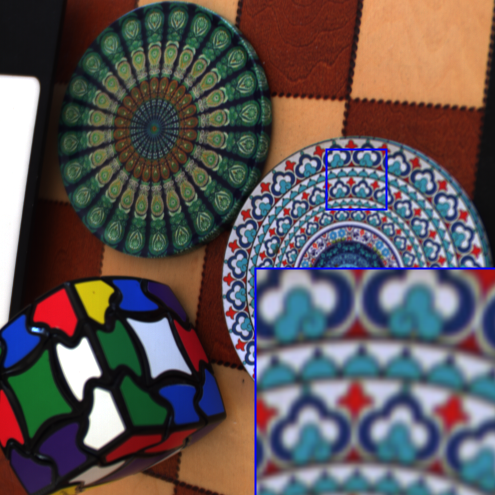}\label{fig:filter1}}
    \hfil
       \subfloat{\includegraphics[width=0.245\textwidth]{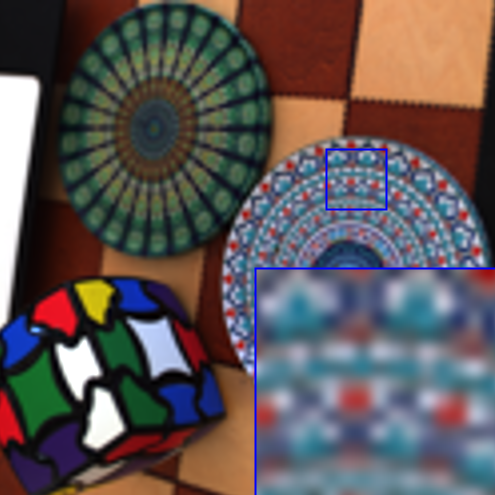}\label{fig:filter2}} 
            \hfil
     \subfloat{\includegraphics[width=0.245\textwidth]{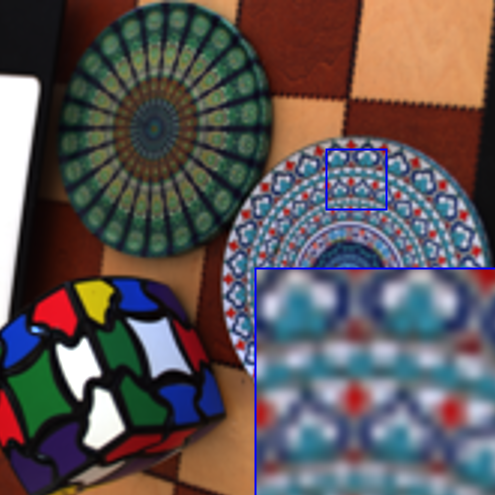}\label{fig:filter3}} 
        \hfil
      \subfloat{\includegraphics[width=0.245\textwidth]{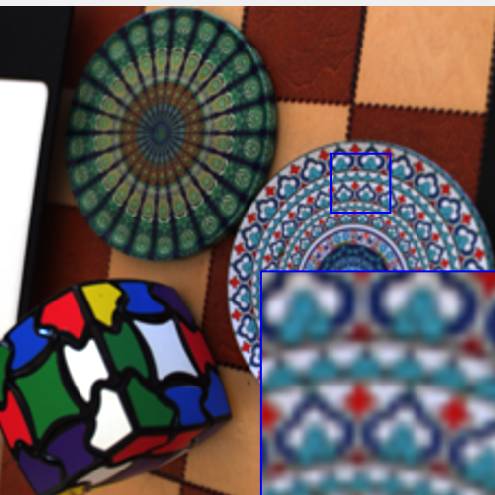}\label{fig:filter3}}  \\
    \caption{Visual Results on the Spectral World dataset for scaling factor $\times 4$. Spectral bands $80$, $45$, and $20$ are used as the R, G, and B channels of a color image for this visualization. We compare the baseline method LIIF~\cite{implicit} and its test-time trained version by our method.}
    \label{fig:results1}
\end{figure*}

\begin{figure*}[!tb]
    \centering
    \subfloat[Ground Truth]{\includegraphics[width=0.245\textwidth]{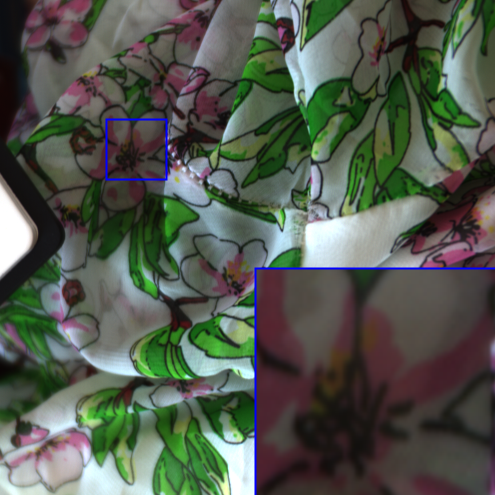}\label{fig:filter1}}
    \hfil
       \subfloat[Bilinear Interpolation]{\includegraphics[width=0.245\textwidth]{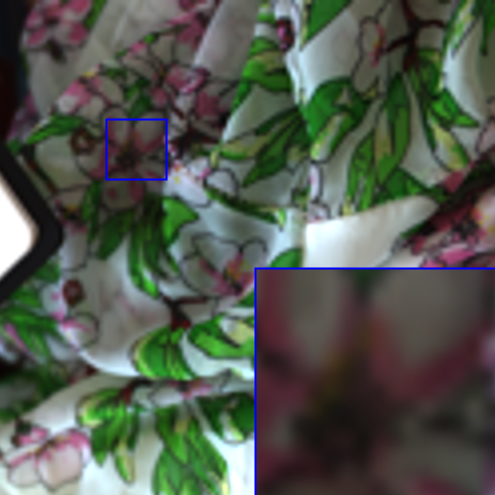}\label{fig:filter2}} 
            \hfil
     \subfloat[Source Model \cite{Li_2022_WACV}]{\includegraphics[width=0.245\textwidth]{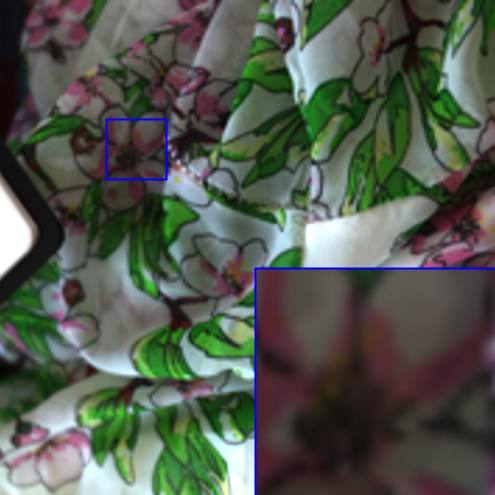}\label{fig:filter3}} 
        \hfil
      \subfloat[Test-time Trained (Ours)]{\includegraphics[width=0.245\textwidth]{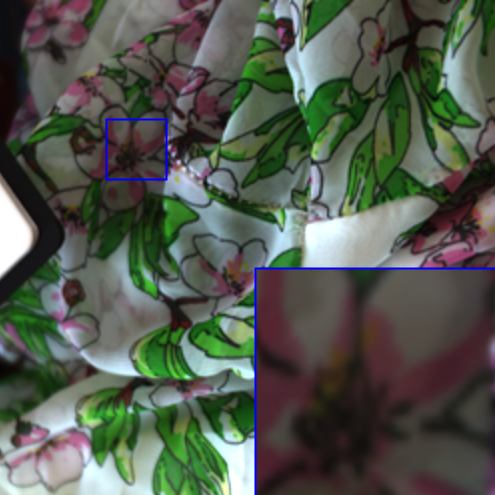}\label{fig:filter3}}  \\ \vspace{-3mm}
    \subfloat{\includegraphics[width=0.245\textwidth]{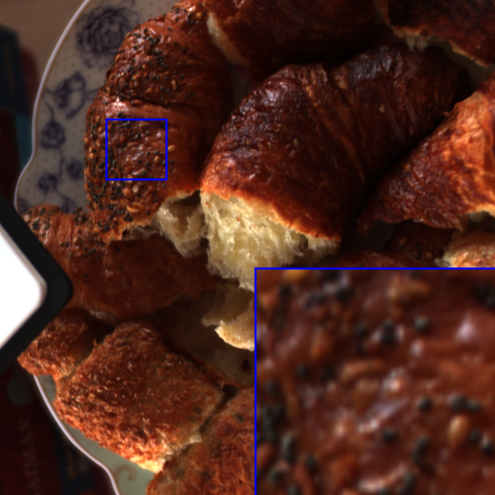}\label{fig:filter1}}
    \hfil
       \subfloat{\includegraphics[width=0.245\textwidth]{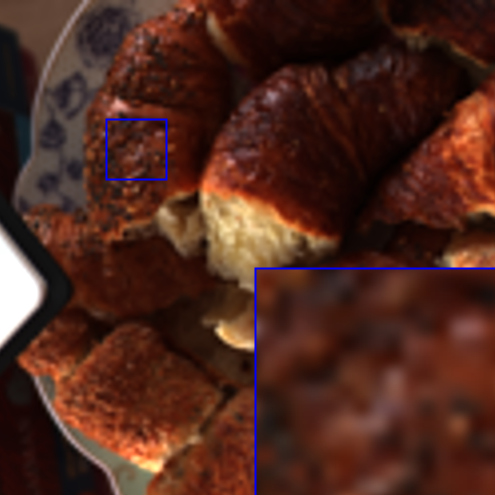}\label{fig:filter2}} 
            \hfil
     \subfloat {\includegraphics[width=0.245\textwidth]{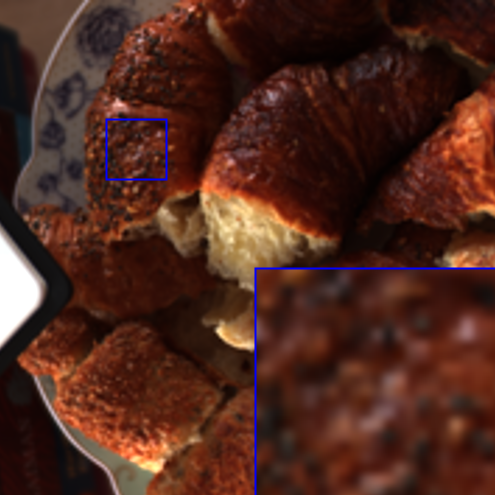}\label{fig:filter3}} 
        \hfil      \subfloat{\includegraphics[width=0.245\textwidth]{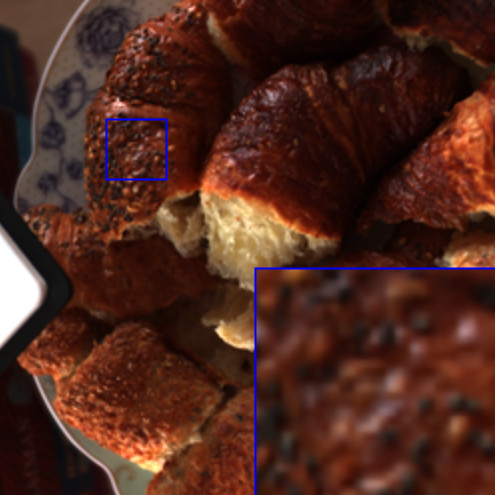}\label{fig:filter3}}  \\ \vspace{-3mm}
        \subfloat{\includegraphics[width=0.245\textwidth]{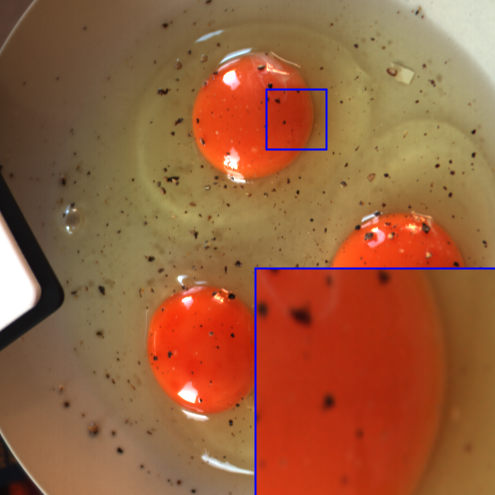}\label{fig:filter1}}
    \hfil
       \subfloat {\includegraphics[width=0.245\textwidth]{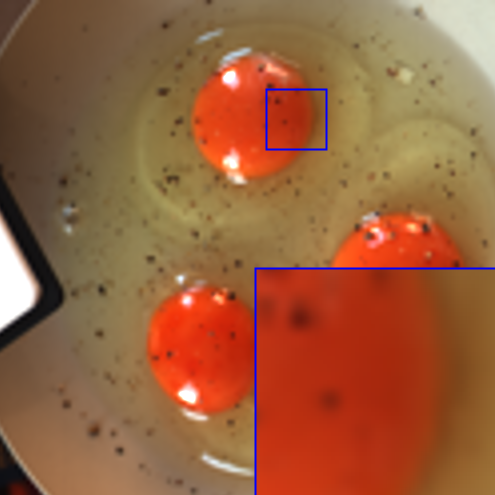}\label{fig:filter2}} 
            \hfil
     \subfloat{\includegraphics[width=0.245\textwidth]{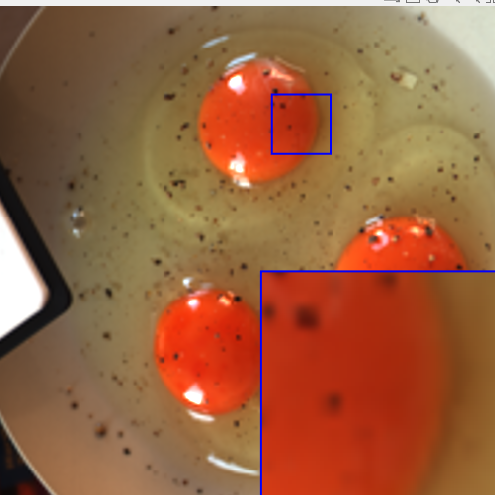}\label{fig:filter3}} 
        \hfil
      \subfloat{\includegraphics[width=0.245\textwidth]{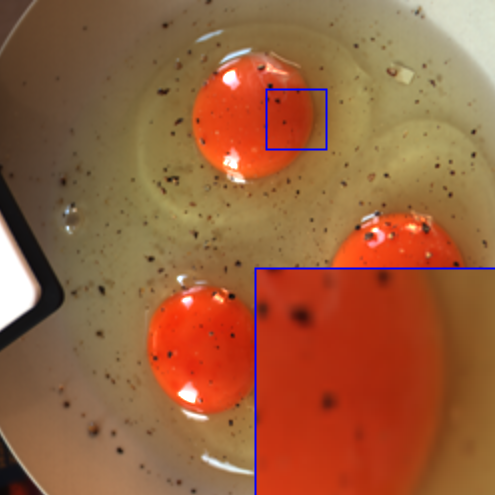}\label{fig:filter3}}  \\
      \vspace{-3mm}
        \subfloat{\includegraphics[width=0.245\textwidth]{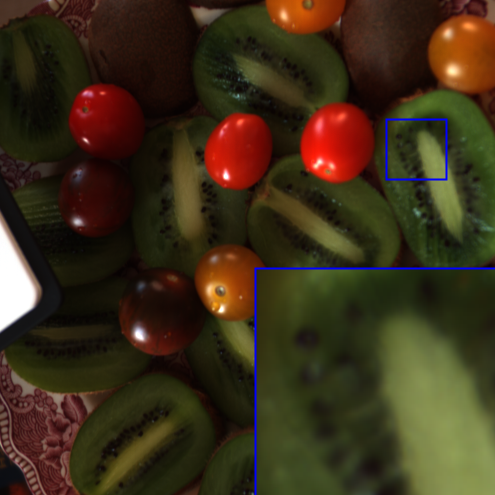}\label{fig:filter1}}
    \hfil
       \subfloat {\includegraphics[width=0.245\textwidth]{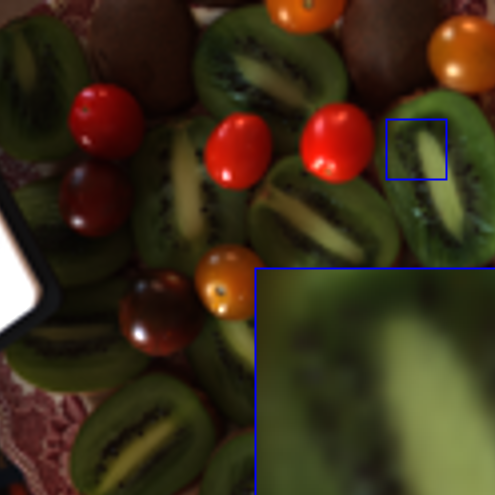}\label{fig:filter2}} 
            \hfil
     \subfloat{\includegraphics[width=0.245\textwidth]{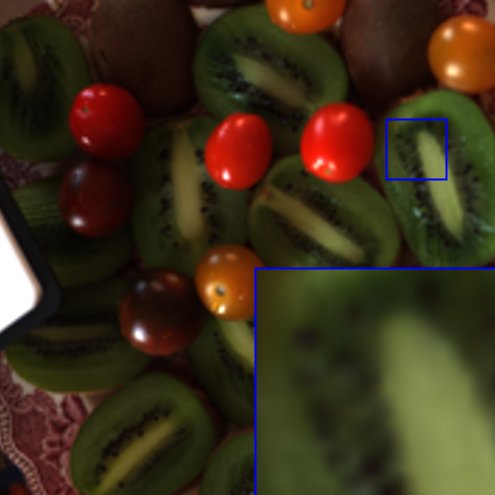}\label{fig:filter3}} 
        \hfil
      \subfloat{\includegraphics[width=0.245\textwidth]{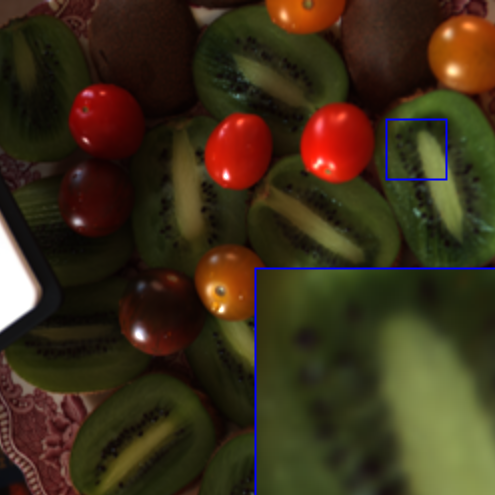}\label{fig:filter3}} \\
    \caption{Visual Results on the Spectral World dataset for scaling factor $\times 4$. Spectral bands $80$, $45$, and $20$ are used as the R, G, and B channels of a color image for this visualization. We compare the baseline method HSIwRGB~\cite{Li_2022_WACV} and its test-time trained version by our method.}
    \label{fig:results2}
\end{figure*}

\begin{figure*}[!bt]
    \centering
    \subfloat[Ground Truth]{\includegraphics[width=0.245\textwidth]{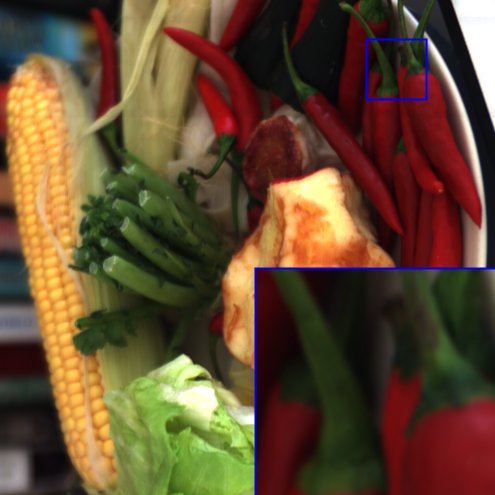}\label{fig:filter1}}
    \hfil
       \subfloat[Bilinear Interpolation]{\includegraphics[width=0.245\textwidth]{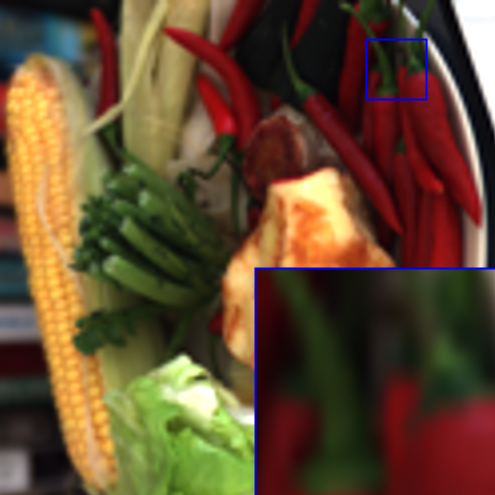}\label{fig:filter2}} 
            \hfil
     \subfloat[Source Model \cite{swinIR}] {\includegraphics[width=0.245\textwidth]{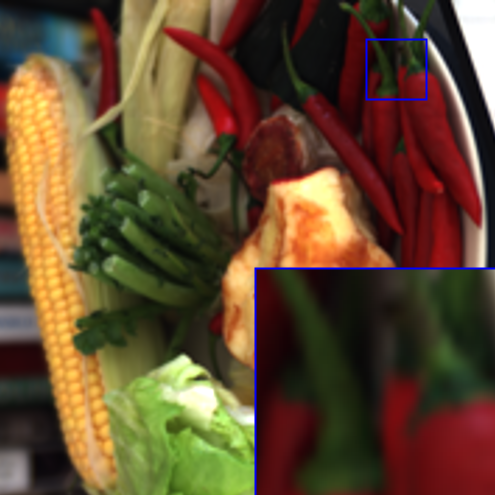}\label{fig:filter3}} 
        \hfil
      \subfloat[Test-time Trained (Ours)]{\includegraphics[width=0.245\textwidth]{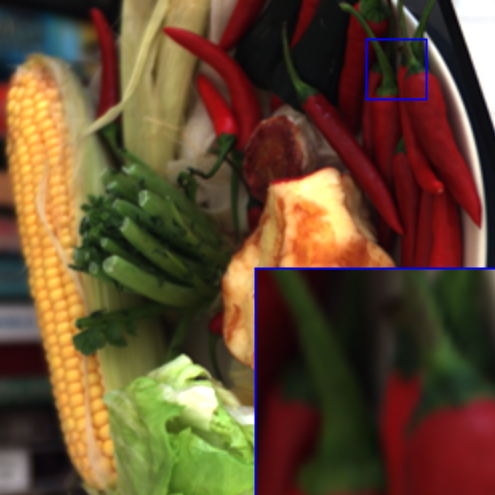}\label{fig:filter3}}  \\ \vspace{-3mm}
    \subfloat{\includegraphics[width=0.245\textwidth]{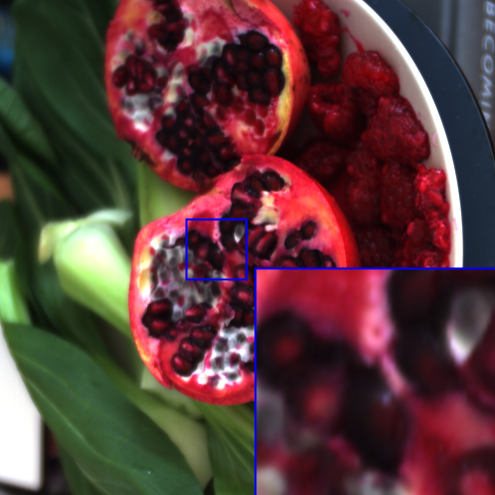}\label{fig:filter1}}
    \hfil
       \subfloat{\includegraphics[width=0.245\textwidth]{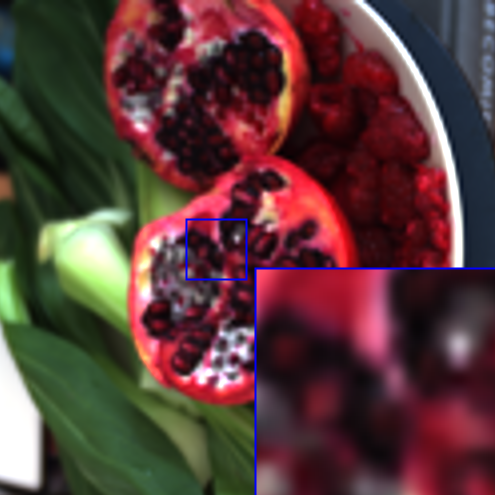}\label{fig:filter2}} 
            \hfil
     \subfloat {\includegraphics[width=0.245\textwidth]{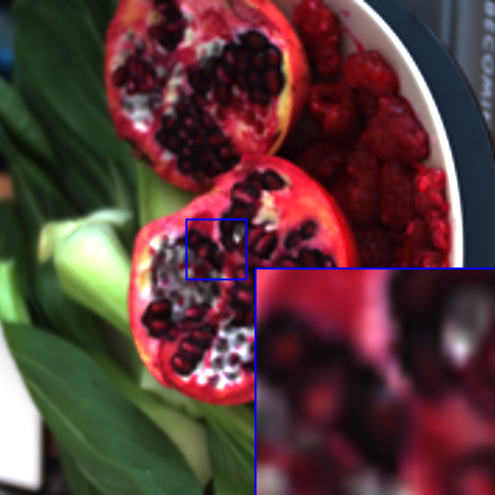}\label{fig:filter3}} 
        \hfil      \subfloat{\includegraphics[width=0.245\textwidth]{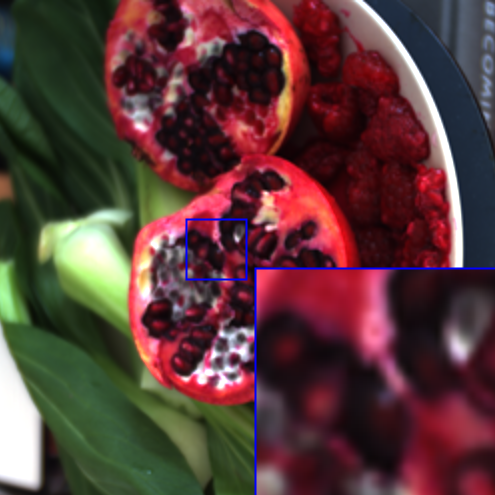}\label{fig:filter3}}  \\ \vspace{-3mm}
        \subfloat{\includegraphics[width=0.245\textwidth]{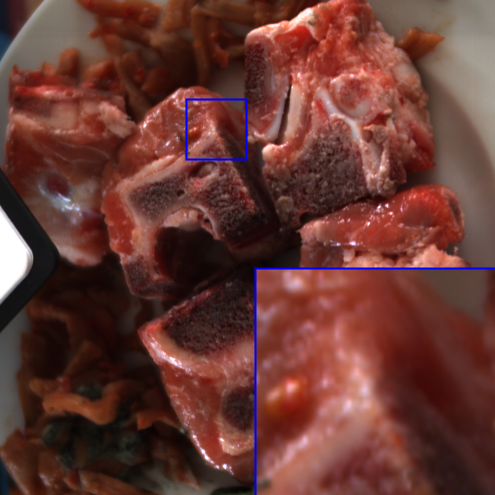}\label{fig:filter1}}
    \hfil
       \subfloat{\includegraphics[width=0.245\textwidth]{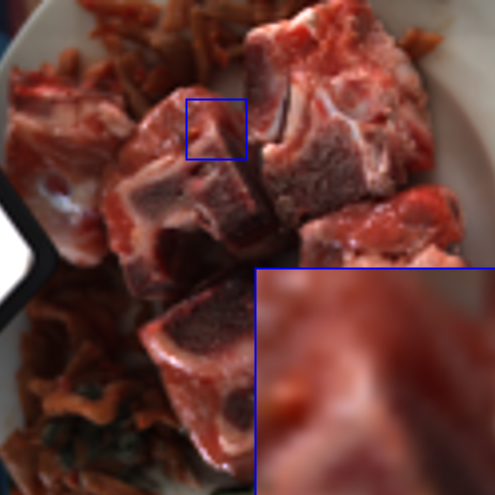}\label{fig:filter2}} 
            \hfil
     \subfloat {\includegraphics[width=0.245\textwidth]{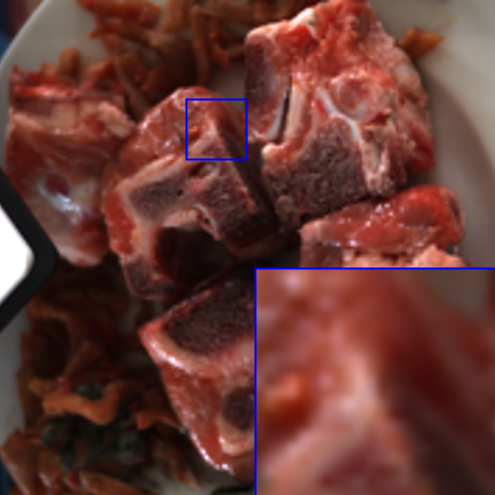}\label{fig:filter3}} 
        \hfil
      \subfloat{\includegraphics[width=0.245\textwidth]{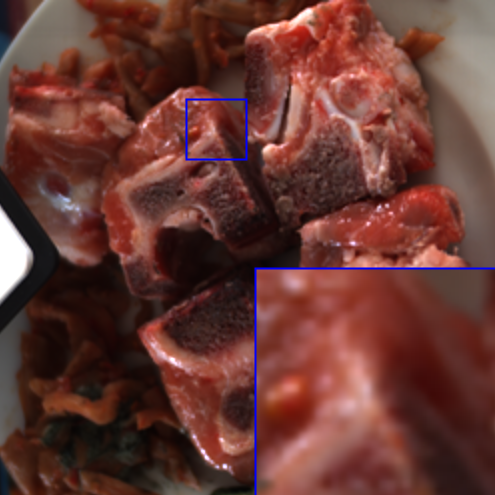}\label{fig:filter3}}  \\
      \vspace{-3mm}
        \subfloat{\includegraphics[width=0.245\textwidth]{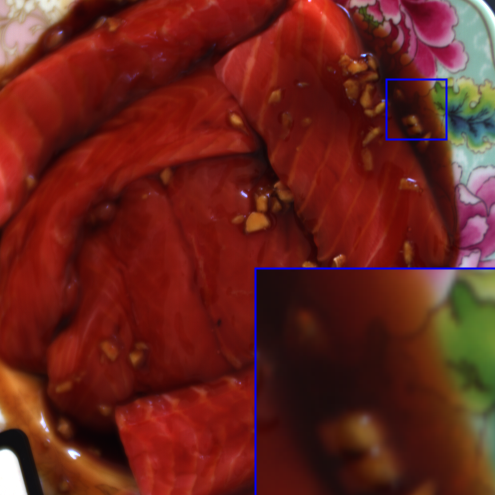}\label{fig:filter1}}
    \hfil
       \subfloat{\includegraphics[width=0.245\textwidth]{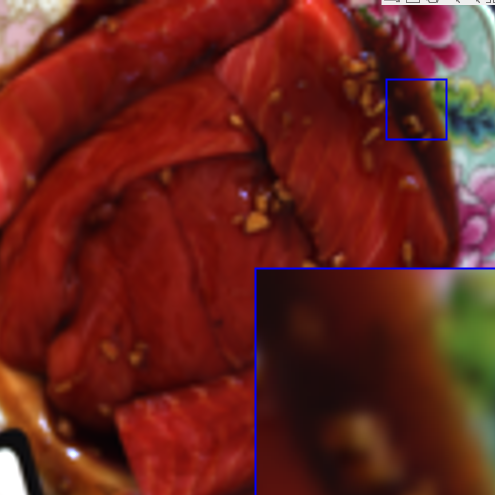}\label{fig:filter2}} 
            \hfil
     \subfloat {\includegraphics[width=0.245\textwidth]{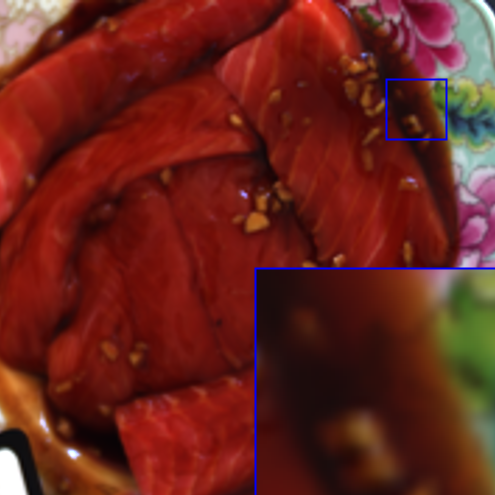}\label{fig:filter3}} 
        \hfil
      \subfloat{\includegraphics[width=0.245\textwidth]{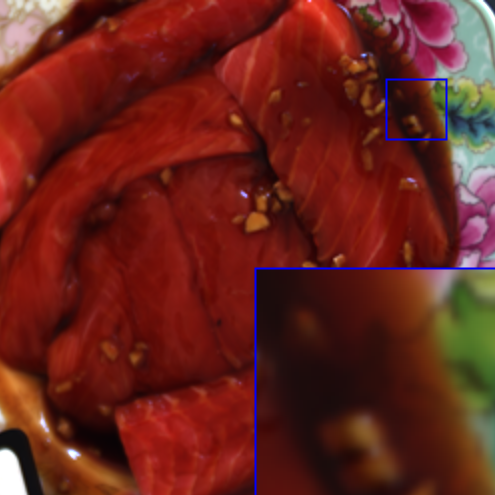}\label{fig:filter3}}  \\
    \caption{Visual Results on the Spectral World dataset for scaling factor $\times 4$. Spectral bands $80$, $45$, and $20$ are used as the R, G, and B channels of a color image for this visualization. We compare the baseline method SwinIR~\cite{swinIR} and its test-time trained version by our method.}
    \label{fig:results3}
\end{figure*}

First, we find that our self-training with loss in Eq.~\ref{eq:pseudo} with the synthesized pseudo samples can indeed enable test-time learning for HSI SR. It actually improves the performance of the source-only pre-trained model by a significant margin. The improvement is because the generated LR-HR relationship is accurate so that useful knowledge of HSI SR can be learned. Furthermore, the generated LrHSI is very similar to the actual LrHSI so that the updated model can be well `generalized' to the real test sample. 

Second, the weights-averaged teacher-student learning method further improves performance. This improvement is due to the use of the more accurate weight-averaged prediction as the pseudo-label target; and 2) the mean teacher prediction encodes the information from models in past iterations and is, therefore, less likely to suffer from catastrophic forgetting over the learning iterations.

\begin{table}[!tb]
  \centering
  \small
  \begin{tabular}{lcccccccccccccc}
\toprule
Methods & CAVE & Harvard & Spectral World\\ \midrule
Color Jittering & 38.9543 & 39.0024 & 40.0231\\
Image Rotation &  38.9421 & 39.0964 & 40.0753\\
Mixup \cite{zhang2018mixup} & 37.2530 & 37.4311 & 39.2382\\
Cutblur Mixup \cite{data:augmentation:SR:cvpr20} & 39.0013 & 39.1256 & 40.1042\\
Spectral Mixup    &  \textbf{39.2376} &  \textbf{39.4467} & \textbf{40.2083} \\ 
\bottomrule 
\end{tabular} 
\caption{Ablation study for \emph{Spectral Mixup}} 
\label{tab:ablation:mixup}
    \end{table}

Third, the results in Table~\ref{tab:2} and Table~\ref{tab:3} also show that the proposed data augmentation method Spectral Mixup can also improve performance. While the improvement is not as large as the other two contributions, it is consistent across different datasets and different scaling factors. Therefore, it can be used as a simple plug-in for HSI SR methods.  We also compare our method with other popular data augmentation methods including color jittering, image rotation,   \emph{mixup}~\cite{zhang2018mixup}, and Cutblur Mixup \cite{data:augmentation:SR:cvpr20}. The results in Table~\ref{tab:ablation:mixup} show that other methods do not provide as much benefit as our method. The key reason is that our data augmentation better preserves the detailed structure of the HSIs while introducing variations to the data. 
 The spatial mixing method \cite{zhang2018mixup} blends data from two images. It may break detailed structures that are important for SR tasks. The CutblurMixup method learns \emph{where to perform the SR} and is found helpful for RGB SR \cite{data:augmentation:SR:cvpr20}.  \emph{Spectral Mixup} creates virtual examples by using convex combinations of spectral bands of the same image which avoids breaking image structures and preserves simple linear behavior in-between spectral bands. This makes it more suitable for HSI tasks. The other two general data augmentation methods Color Jitterring and Image Rotation are slightly helpful as well but not as effective as Spectral Mixup. 
Spectral Mixup trains a neural network on convex combinations of HSI channels and their labels. By doing so, it regularizes the neural network to favor simple linear behavior in-between HSI channels to alleviate the over-fitting issues.

\begin{table}[!tb]
  \centering
  \setlength\tabcolsep{1.8pt}
  \small
  \begin{tabular}{cccccccc}
\toprule
 \multicolumn{1}{c} {Method} & \multicolumn{1} {c} {CAVE} & \multicolumn{1} {c} {Harvard} &\multicolumn{1} {c} {Spectral World}\\ \toprule
 Source (LIIF~\cite{implicit})   &  37.7645 &  37.9317 & 38.9642 \\ 
Ours (LIIF~\cite{implicit})   &  \textbf{39.2376} &  \textbf{39.4467} & \textbf{40.2083} \\ 
\midrule
 Source (HSIwRGB~\cite{Li_2022_WACV})   &   43.3242 &  41.0589 & 41.2375 \\ 
Ours (HSwRGB~\cite{Li_2022_WACV})   &  \textbf{44.2489} &  \textbf{42.0236} & \textbf{42.0874} \\ 
\midrule
 Source (SwinIR~\cite{swinIR})   &  42.0143 & 41.2724  & 40.8572 \\ 
Ours (SwinIR)   &  \textbf{43.3564} &  \textbf{42.0059} & \textbf{41.9532} \\ 
\bottomrule
\end{tabular} 
\caption{Results (PSNR) for scaling factor $\times 4$.  }
  \label{tab:5} 
\end{table}

Fourth, we also compare our method to existing test-time training methods. While test-time training and adaptation have received quite some attention in the past years, almost all research efforts have been made on the classification task (as shown in this survey paper \cite{Liang2023ACS}). Therefore, we need to adapt classification methods for this comparison. Most previous methods employ the idea of \emph{learning with categorical pseudo-labels} via \emph{cross-entropy minimization} such as TENT~\cite{wang2020tent} and CoTTA \cite{Wang_2022_CVPR}. Those methods do not work for our task as directly using the input LR image and its HR prediction (their pseudo-labels) to re-train our SR model leads to zero model update. Our method can be understood as a novel extension of pseudo-label-based methods for HSI SR. As such, for comparison, we choose to extend three recent methods in the other family of methods that self-trains models with an auxiliary task: image rotation in TTT~\cite{sun19ttt}, MAE in TTT-MAE~\cite{gandelsman2022testtime}, and contrasting prototypes in TTAPS~\cite{Bartler2022TTAPSTA}. We use the same backbone for the three methods as for our method. As shown in Table~\ref{tab:2} and Table~\ref{tab:3}, they do not yield good results for HSI SR. Updating the model parameters via using a self-supervised auxiliary task can corrupt the model weights. This is especially true when the auxiliary task and the primary task are not closely related, and when the primary tasks are very delicate. This is exactly the case here when those image classification methods are applied to HSI SR. 

Finally, we evaluate how our self-training method works with different network backbones. As stated in the method section, we choose LIIF~\cite{implicit} as the baseline because we would like to have the flexibility of arbitrary-scale upsampling -- one model for all scaling factors. That is also one of the reasons why our baseline model performs worse than the SOTA methods such as \cite{Li_2022_WACV} for a fixed scaling factor. We here extend our evaluation to include the SOTA HSI SR method \cite{Li_2022_WACV} and the popular Transformer-based approach SwinIR \cite{swinIR}. They are all trained with the single-channel processing setup used by our method. The results in Table \ref{tab:5} show that our method improves the performance of all baselines consistently and considerably. Stronger baseline models will lead to higher-quality virtual training samples for our test-time training, which will lead to better models, meaning our method can also benefit from using stronger baseline models. This also means that our proposed test-time training method can be combined with any HSI SR method for further improvement.

\textbf{Visual Results}. In this section, we show visual results of different scenes/objects to compare the performance of our method with the corresponding baseline methods. Specifically, we show visual results of the baseline method LIIF~\cite{implicit} and its test-time trained version by our method in Figure~\ref{fig:results1}, we show visual results of the baseline method HSIwRGB~\cite{implicit} and its test-time trained version by our method in Figure~\ref{fig:results2}, and we show visual results of the baseline method SwinIR~\cite{implicit} and its test-time trained version by our method in Figure~\ref{fig:results3}. Images are all selected from our Spectral World dataset, where HS channel $80$, $45$, and $20$ are selected and used as channel `R', `G', and `B' of an RGB image for visualization. There is no other color processing step. The ground-truth images and the results via simple Bilinear interpolation are also shown for reference. 
The visual results of a diverse range of images in the three figures further confirm that our test-time training method can improve the quality of the SR results quite significantly. Subtle structures and textures (such as the calyx of the strawberry in Figure~\ref{fig:results1}, the distinctive texture of the Croissant in Figure~\ref{fig:results2}, the boundary between egg white and egg yolk in Figure~\ref{fig:results2}, and the structure of the bones and meat in Figure~\ref{fig:results3}) can be better recovered now by test-time training with our method. 

\section{Conclusion}
In this paper, we have presented a novel test-time training method for hyperspectral image (HSI) super-resolution (SR). Three contributions have been made. First, a novel extension is made to the standard self-training method to generate effective LR-HR HSI training samples for a given test LrHSI and to avoid model divergence during test-time training. Second, a simple, yet effective neural network architecture has been introduced for HSI SR and its test-time training. Third, a new data augmentation method Spectral Mixup is developed to generate more training samples during test-time training. We have also proposed a new HSI dataset with a diverse set of interesting objects. Experiments on three datasets show that our test-time training method can significantly improve the performance of a pre-trained HSI SR model. The method is simple and easy to use. Experimental results show that it can be combined with different HSI SR methods for further improvement.

\bibliographystyle{IEEEtran}
\bibliography{main}

\ifpeerreview \else


\begin{IEEEbiography}[{\includegraphics[width=1in,clip,keepaspectratio]{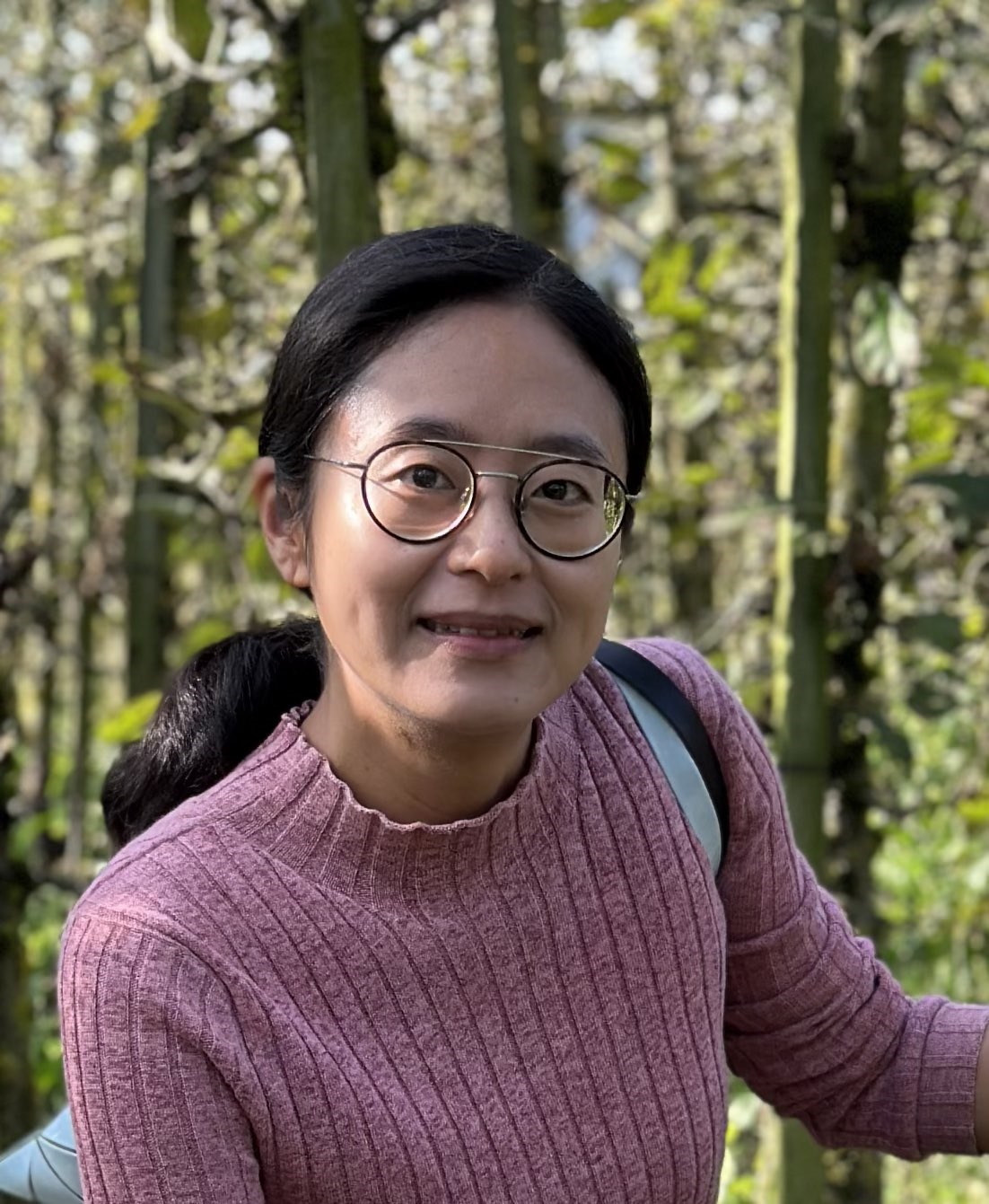}}]{Ke Li}
has received her PhD degree in computer vision from ETH Zurich, in 2024. Before that, she received her Bachelor's degree in Clinical Medicine from Tongji Medical School, Huazhong University of Science and Technology, China, in 2011. She obtained her first Master's degree in gene expression analysis from the University of Basel, in 2015, and her second Master's degree in Biostatistics from the University of Zurich, in 2018. Her research interests lie in Hyperspectral Imaging, Hyperspectral Image Super-resolution, Learning with Auxiliary Tasks, and Domain Adaptation.
\end{IEEEbiography}

\begin{IEEEbiography}[{\includegraphics[width=1in,clip,keepaspectratio]{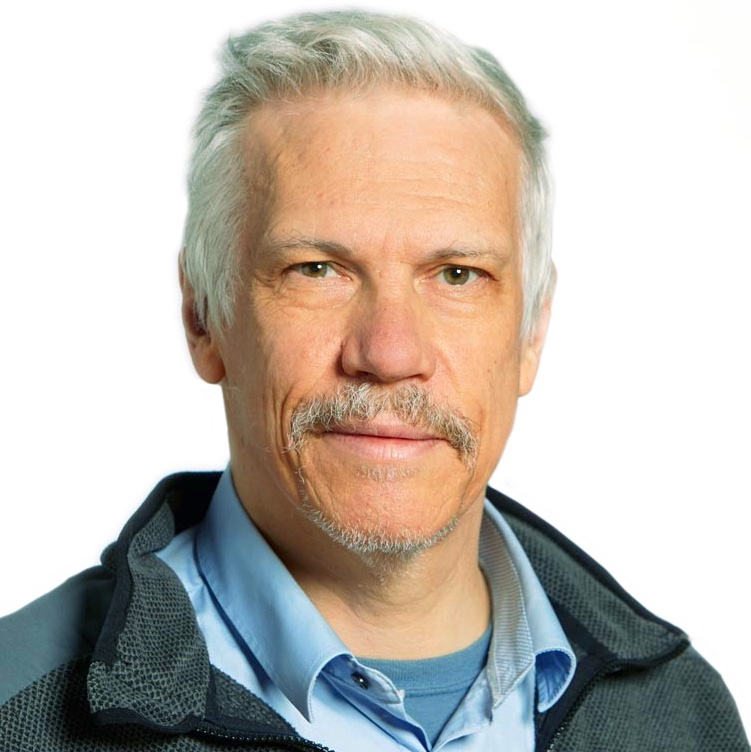}}]{Luc Van Gool}
is a full professor for Computer Vision at ETH Z\"urich, the KU Leuven and INSAIT. He leads research and/or teaches at all three institutions. He has authored over 900 papers. He has been a program committee member of several major computer vision conferences (e.g.\ Program Chair ICCV’05, Beijing, General Chair of ICCV’11, Barcelona, and of ECCV’14, Z\"urich). His main interests include 3D reconstruction and modeling, object recognition, and autonomous driving. He received several best paper awards (e.g.\ David Marr Prize ’98, Best Paper CVPR’07). He received the Koenderink Award in 2016 and the ``Distinguished Researcher'' nomination by the IEEE Computer Society in 2017. In 2015 he also received the 5-yearly Excellence Prize by the Flemish Fund for Scientific Research. He was the holder of an ERC Advanced Grant (VarCity). Currently, he leads computer vision research for autonomous driving in the context of the Toyota TRACE labs at ETH and in Leuven, and has an extensive collaboration with Huawei on the topic of image and video enhancement.
\end{IEEEbiography}

\begin{IEEEbiography}[{\includegraphics[width=1in,clip,keepaspectratio]{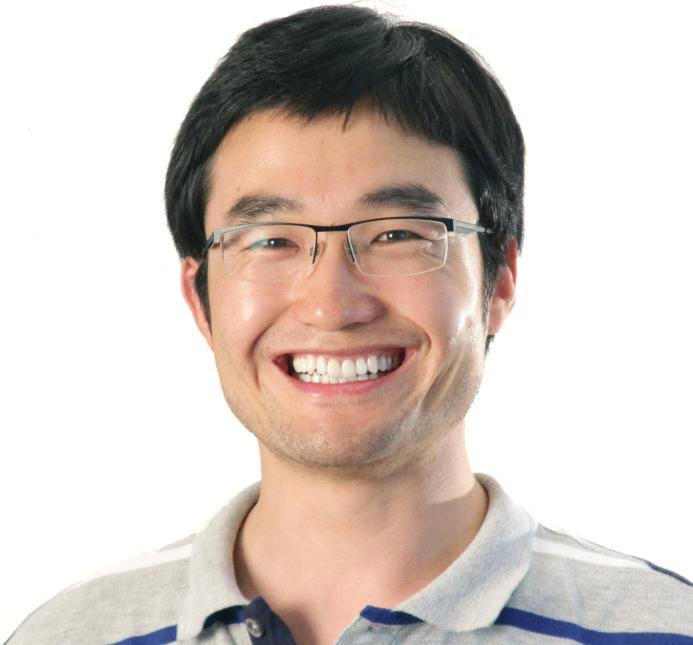}}]{Dengxin Dai}
has received his PhD degree in computer vision from ETH Z\"urich, in 2016. He is now Research Director with Huawei Z\"urich Research Center. Before that, he was a senior group leader with the Max Planck Institute for Informatics and a senior scientist with ETH Z\"urich. He is a member of the ELLIS Society. He has organized multiple well-received workshops, is currently an associate editor of the International Journal of Computer Vision, and has been area chair of multiple vision conferences including CVPR, ECCV, and ICRA. He received the Golden Owl Award with ETH Z\"urich in 2021 for his exceptional teaching and received the German Pattern Recognition Award in 2022 for his outstanding scientific contribution in the area of scalable and robust visual perception.

\end{IEEEbiography}


\fi

\end{document}